\documentclass[reqno, 12pt]{article}
\pdfoutput=1
\usepackage{amssymb}
\usepackage[utf8]{inputenc}
\usepackage{amsmath}
\usepackage{amsfonts}
\usepackage{pdfpages}
\usepackage{graphicx}
\usepackage[a4paper, total={6in, 10in}]{geometry}
\usepackage{amssymb}
\usepackage{amsmath}
\usepackage{authblk}
\usepackage{setspace}
\usepackage{array}
\usepackage{xcolor}
\usepackage{amsthm}
\usepackage{multicol}
\newtheorem{theorem}{Theorem}
\newtheorem{lemma}{Lemma}

\newtheorem*{proposition}{Proposition}
\newtheorem{remark}{Remark}
\usepackage{hyperref}
\hypersetup{
    colorlinks=true,
    linkcolor=blue,
    filecolor=magenta,      
    urlcolor=cyan
    }
\usepackage{tensor}
\title{On the limits of neural network explainability via descrambling}
\author[1]{Shashank Sule\thanks{ssule25@umd.edu}}
\author[2]{Richard G. Spencer\thanks{spencerri@mail.nih.gov}}
\author[1]{Wojciech Czaja\thanks{wojtek@math.umd.edu}}
\affil[1]{\small{Department of Mathematics, University of Maryland, College Park, MD 20742, USA}}
\affil[2]{\small{National Institute on Aging, National Institutes of Health }
}
\begin{document}
\maketitle
\vspace{-0.3in}
\begin{abstract}
We characterize the exact solutions to \emph{neural network descrambling}--a mathematical model for explaining the fully connected layers of trained neural networks (NNs). By reformulating the problem to the minimization of the Brockett function arising in graph matching and complexity theory we show that the principal components of the hidden layer preactivations can be characterized as the optimal ``explainers" or \emph{descramblers} for the layer weights, leading to \emph{descrambled} weight matrices. We show that in typical deep learning contexts these descramblers take diverse and interesting forms including (1) matching largest principal components with the lowest frequency modes of the Fourier basis for isotropic hidden data, (2) discovering the semantic development in two-layer linear NNs for signal recovery problems, and (3) explaining CNNs by optimally permuting the neurons. Our numerical experiments indicate that the eigendecompositions of the hidden layer data--now understood as the descramblers--can also reveal the layer's underlying transformation. These results illustrate that the SVD is more directly related to the explainability of NNs than previously thought and offers a promising avenue for discovering interpretable motifs for the hidden action of NNs, especially in contexts of operator learning or physics-informed NNs, where the input/output data has limited human readability. 
\end{abstract}
\section{Introduction}
\label{sec:intro}
Explaining or interpreting the mapping between inputs and outputs of trained neural networks (NNs) remains one of the major unsolved problems in machine learning and is the focus of significant research effort \cite{yosinski2015understanding, simonyan2014deep}. Recently these efforts focused on associating individual components of NNs (such as layers or groups of neurons) with mathematical transformations (such as the Fourier transform \cite{nanda2023progress} or wavelet shrinkage \cite{kadkhodaie2023generalization}). Such achievements in explaining NN weights have relied on sophisticated strategies for processing the given NN in conjunction with its training data. Moreover, these explanations have required considerable  experimentation and as such, the existence of a general method for identifying the deterministic transformations within the hidden layers of NNs is still an elusive goal in machine learning. An intriguing solution towards this goal was recently proposed in \cite{amey2021neural} which introduced \emph{descrambling transformations}, the subject of this paper. 

\subsection{Neural Network Descrambling}

NN descrambling is a mathematical model for explaining trained neural networks. Specifically, the goal of NN descrambling is to create human/expert readable visualizations of the trained weights $W_k$. To this end, let $f: \mathbb{R}^{n_0} \to \mathbb{R}^{n_L}$ be a \emph{bias-free} $L$-layer neural network given by the alternating compositions of linearities and nonlinearities: 
\begin{align}
    f(x) := W_L \phi  \cdots  \phi W_1 x. \label{eq:nn}
\end{align}
It is assumed that the set of NN weights $\{W_i\}_{i=1}^{L}$ is obtained after optimizing some loss function on the input-output data $X \in \mathbb{R}^{n_0 \times N}$ and $Y \in \mathbb{R}^{n_L \times N}$. Here each column of $X$ (resp. $Y$) represents an input (resp. output) point. The weights $\{W_i\}_{i=1}^{L}$ are considered uninterpretable or \emph{scrambled}. In NN descrambling, each weight matrix $W_k$ is brought into a readable or \emph{descrambled} form $\widehat{P}W_k$, through multiplication by a matrix $\widehat{P}$--the \emph{descrambler}. The descrambler is constructed as follows: in the chain of compositions \eqref{eq:nn} we insert the identity matrix factorized as $I = P^{-1}P$ in the $k$th layer so that the network's output $F$ remains unchanged after propagating $X$ column-wise: 
\begin{equation}
    F = W_L \cdots \phi P^{-1} PW_k \cdots W_1 X = W_L \cdots \phi P^{-1} P f_k(X),\label{eq: wiretap}
\end{equation}
where $f_k(X) = W_k \phi \cdots \phi W_1 X \in \mathbb{R}^{n_{k} \times N}$ is the matrix of \emph{preactivations}
%
given by the intermediate data in the network at the $k$th layer. The matrix $P$ thus acts on the preactivations; in NN descrambling we choose $P$ which transforms the preactivation data optimally for minimizing a well-chosen explainability loss function $\eta$ termed over a specified matrix group $G$:
\begin{align}
    \widehat{P}\,(\eta, G, k, X, N) &= \underset{P \in G}{\text{argmin}}\: \eta\,(Pf_k(X)). \label{eq: descrambling equation}
\end{align}
An important choice for $\eta$ proposed in \cite{amey2021neural} was the smoothness descrambling loss, where the preactivations $f_k(X)$ are rearranged to be optimally smooth: 
\begin{align}
    \eta_{SC}\,(P f_k(X)) = \|D P f_k(X) \|_{F}^{2}. \label{eq: smoothness function}
\end{align}
Here $D$ is taken to be either a spectral or finite-difference second order differentiation matrix \cite{trefethen2000spectral}, so minimizing $\eta_{SC}$ corresponds to choosing the orthogonal change of basis such that the discrete second derivative of the signal $Pf_k(X)$ is minimized. The motivation behind this choice of NN interpretation is that while in signal/audio processing contexts, the incoming and outgoing data is usually smooth and has an intelligible time-ordered structure, the intermediate data often loses this structure. As such, the role of the descrambler is to change basis so that the intermediate signal is itself smooth and ordered with respect to the indices of the output or input dimension. Using the formulation \eqref{eq: descrambling equation} the descramblers $\widehat{P}_{SC}$ are obtained as the minimizers of $\eta_{SC}$ over the orthogonal matrices $O(n_k)$, where $n_k$ is the number of neurons in the $k$th layer: 
\begin{align}
     \widehat{P}_{SC}(k,X,N) := \widehat{P}(\eta_{SC}, O(n_k), k, X, N) = \underset{P^{\top}P = I}{\text{argmin}} \: \|DPf_k(X)\|_{F}^{2}.  \label{eq: smoothness criterion}
\end{align}

The optimization problem \eqref{eq: smoothness criterion} is an example of a quadratically constrained quadra\-tic program (QCQP). We will demonstrate that \eqref{eq: smoothness criterion} admits several global minima. Given that we identify $\widehat{P}_{SC}(k,X,N)$ with \emph{explanations} of the weights $W_k$, the non-uniqueness of the minimizers means that there may be \emph{multiple} explanations of the $k$th layer weights. To select a particular explanation, the authors in \cite{amey2021neural} suggested a gradient-based optimization technique based on the Cayley transform such that the initial guess starts from ``no explanation", $P = I$, and then iteratively minimizes the explainability loss. 

Smoothness descrambling was used to explain the two-layer network \emph{DEERNet} \cite{worswick2018deep} trained to solve an integral equation arising in double electron-electron resonance (DEER). After descrambling, the first layer of the network was deemed to represent a cubic time-distance conversion in the DEER model and the second layer was judged to represent projection in the basis of Chebyshev polynomials (see Figs. \ref{fig: 2panel} \& \ref{fig: 6panel_chebyshev} or \cite[Figures 2,4]{amey2021neural}). In an ablation study, these layers were replaced by the identified mathematical transformations such as notch filtering, leading to a deterministic method for processing DEER data and removing the need for an NN entirely.

\subsection{Goals of this work and summary of results}

We aim to give a theoretical account for how descramblers 
explain the weights $W_k$. Specifically, our goal is not to reverse engineer networks but to characterize the $\widehat{P}_{SC}(k,X,N)$ which enable the aforementioned reverse-engineering. Moreover, we will suggest an alternative method to find the minimizers of \eqref{eq: smoothness criterion} based on an eigendecomposition and show that it may provide clearer explanations of the network weights. We also restrict ourselves to cases where we can take the limit of the descramblers as $N \to \infty$. Our results can be summarized as follows: 
\smallskip

    \noindent {\bf 1.} For NN descrambling using $\eta_{SC}$ with a general $n_k \times n_k$ matrix $D$, there exists a sequence of descramblers $\widehat{P}(k,X,N)$ that converges almost surely to $TU^{\top}$ as the number of points $N \to \infty$, where $T$ are the right singular vectors of $D$ arranged in ascending order of the singular values and $U$ is the matrix of the principal components of the autocorrelation matrix of the preactivations $S = f_{k}(X)$ arranged in \emph{descending} order of the singular values of $S$ (Theorem \ref{thm: main theorem}). 
    
    \noindent {\bf 2.} For smoothness descrambling where $D$ is a second order differentiation stencil, we characterize descramblers for cases where (1) data $X$ is described by isotropic distributions or noisy measurements of signals, (2) NN is trained under the Saxe-Mclelland-Ganguli (SMG) model \cite{saxe2014exact} for a linear signal estimation problem and (3) the network is convolutional. Although these represent highly idealized situations, we note that they nonetheless occur quite commonly in deep learning workflows. 
    
    \noindent {\bf 3.} We revisit the empirical results from \cite{amey2021neural}, showing that we can recover the notch filter in the first layer of DEERNet and an orthogonal polynomial basis in the second layer via an eigendecomposition (instead of solving the problem \eqref{eq: smoothness criterion} via an optimizer). Moreover, we illustrate that although there can be \emph{multiple} ways/descramblers to explain the NN corresponding to multiple minimizers of the problem \eqref{eq: smoothness criterion}, the explanations provided by them are nearly the same.

\section{Results}
\label{sec:results}
The descramblers \eqref{eq: smoothness criterion} for the $k$th layer and data $X$ with $N$ i.i.d. columns are denoted by $\widehat{P}_{SC}(k,X,N)$. We will show that these matrices admit almost sure limits as $N \to \infty$ for a variety of choices of distributions for $X$ and layer number $k$. To characterize such minimizers, first we state and prove a general result which will relate the smoothness objective function \eqref{eq: smoothness function} to the Brockett cost function widely applied in graph matching, complexity theory, and optimization on matrix manifolds \cite{brockett1989least, absil2008optimization}: 

\begin{lemma}
\label{lem:brockett}
    Fix $S \in \mathbb{R}^{d \times n}$ and $A \in \mathbb{R}^{d \times d}$ such that $A^{\top} = A$. Define $\widehat{P}$ to be the solution to the following minimization problem over the orthogonal group $O(d)$:
    \begin{align}
        \widehat{P} \in \underset{P \in O(d)}{\text{argmin}} \, {\textsf Tr}(P^{\top}A P SS^{\top}). \label{eq:brockett function}
    \end{align}
    Let $A = T \Omega T^{\top}$ be an eigendecomposition of $A$ sorted in ascending order of the eigenvalues where $\Omega$ is the diagonal matrix of eigenvalues of A given by $\{\omega_{i}\}_{i=1}^{d}$. Then $\widehat{P}$ is a minimizer of \eqref{eq:brockett function} if there exists a singular value decomposition of $S = U \Sigma V^{\top}$ such that $\widehat{P} = TU^{\top}$. 
\end{lemma}

\begin{remark}
    Note that if we set $S = f_k(X) \subseteq \mathbb{R}^{n_k \times N}$ where $f_k(X)$ are the $k$th layer preactivations then the smoothness descrambling functional $\eta_{SC}(Pf_k(X))$ is given by 
    \begin{align}
       \eta_{SC}(Pf_k(X)) := \|DPS\|^{2}_{F} = {\textsf Tr}(S^{\top}P^{\top}D^{\top}DPS) = {\textsf Tr}(P^{\top}D^{\top}DPSS^{\top}). \label{eq: brockett to descrambling}
    \end{align}
    This is exactly the objective function in \eqref{eq:brockett function} if we set $A = D^{\top}D$. In general for $A,B$ symmetric, the function 
    ${\textsf Tr}(P^{\top}AP B)$ is the Brockett cost function. Remarkably, NN descrambling connects the Brockett function and NN explainability. Lemma \ref{lem:brockett} has also been proved in more general settings using Lagrange multipliers and interlacing inequalities. For completeness, below we present another proof characterizing the minimizers on the orthogonal manifold by relaxation to a generalized eigenvalue problem. 
 \end{remark}

\begin{proof}
    First, using the cyclic property of traces, we get 
    \begin{align}
        F(P) := {\textsf Tr}(P^{\top}A P SS^{\top}) = {\textsf Tr}(S^{\top} P^{\top}A P S).
    \end{align}
    Let $S = U\Theta V^{\top}$ be an SVD of $S$. Then, since $VV^{\top} = I$, we have 
    \begin{align}
        {\textsf Tr}(S^{\top} P^{\top}A P S) = {\textsf Tr}(S^{\top} P^{\top}A P S VV^{\top}) = {\textsf Tr}(V^{\top} S^{\top} P^{\top}A P S V).
    \end{align}
    Now consider the change of variable $Y = PSV$. Then we can reformulate $F(P)$ in terms of $Y$ as follows: 
    \begin{align}
        F(P) = {\textsf Tr}(Y^{\top}A Y) =: G(Y),
    \end{align}
    where $Y \in \mathbb{R}^{d \times n}$ and $Y^{\top} Y = \Theta^{\top} \Theta \in \mathbb{R}^{n \times n}$. Thus, the minimum in problem \eqref{eq:brockett function} can be bounded below by 
    \begin{align}
    \underset{P \in O(d)}{\text{min}} \, F(P)  \geq \underset{Y^{\top}Y = \Theta^{\top}\Theta}{\text{min}} \, {\textsf Tr}(Y^{\top}AY). \label{eq:relaxation}
    \end{align}
    Thus we have relaxed \eqref{eq:brockett function} to a generalized eigenvalue problem. To attain the minimum on the RHS of \eqref{eq:relaxation}, we proceed as follows: the first $r := \textsf{rank}(\Theta)$ columns of the minimizers of $G(Y)$ are given by the first $r$ eigenvectors of $A$ renormalized by $\theta_i$ such that $\|y_i\|^{2}_{2} = \theta^{2}_{i}$. Thus the first $r$ columns of $Y$ are given by $\theta_i t_i$ where $t_i$ are the eigenvectors corresponding to the smallest $r$ eigenvalues of $A$ (including multiplicities and negative eigenvalues). The rest of the $n-r$ columns of $Y$ are given by $y_i = 0$. Thus, overall the solution may be written as $\widehat{Y} = T\Theta$. Plugging this into $G$ we get
    \begin{align}
    G(\widehat{Y}) = \sum_{i=1}^{r}\theta_{i}^{2}\omega_{i}^{2}. \label{eq: minimum of the brockett function}
    \end{align}
    Thus, for a fixed ordering of the non-zero singular values of $S$ the minimum of $G$ is attained by such $\widehat{Y}$. To minimize the RHS of 
    \eqref{eq: minimum of the brockett function} over all possible arrangements of the non-zero singular values of $S$, we note the following: since $\omega_i$ necessarily range from smallest to largest, to minimize \eqref{eq: minimum of the brockett function} we must choose $\theta_{i}^2$ to range from largest to smallest due to the rearrangement inequality. Finally, we consider the original problem. Let $\widehat{P}$ be such that $\widehat{Y} = T\Theta = \widehat{P}SV$. Then we may write $S = \widehat{P}^{\top}T\Theta V$. This is a singular value decomposition for $S$, so there exists some $U \in O(d)$ such that $\widehat{P}^{\top}T = U$ and $S = U\Theta V^{\top}$. Therefore $\widehat{P} = TU^{\top}$, where the columns of $T$ (resp. $U$) are arranged in the ascending (resp. descending) order of the singular values.  
\end{proof}

\begin{remark}
\label{rem: uniqueness}
    Note that the solutions to \eqref{eq: minimum of the brockett function} are not unique and depend (at least) on the selection of the singular vectors of $A$ and $S$. 
\end{remark}

 \subsection{Neural network descrambling: general results}
 
Using Lemma \ref{lem:brockett} the large data limits of the matrices $\widehat{P}_{SC}$ in \eqref{eq: smoothness criterion} are characterized in a similar manner to the consistency of principal component analysis: 

 \begin{theorem}
 \label{thm: main theorem}
     Let $\Sigma = \mathbb{E}_{x}[f_{k}(x)f_k(x)^{\top}]$ be the autocorrelation matrix of $f_k(x)$ and assume $\Sigma$ has distinct eigenvalues. Moreover, let $U\Lambda U^{\top} = \Sigma$ be an eigendecomposition of $\Sigma$. There exists a sequence of minimizers $\widehat{P}_{SC}(k,X,N)$ to \eqref{eq: smoothness criterion} such that as $N \to \infty$, $\widehat{P}_{SC}(k,X,N) \to TU^{\top}$ almost surely in the measure induced by the data $x$. 
 \end{theorem}

 \begin{proof}
    Since a factor of $N^{-1}$ does not change the minimizer in \eqref{eq: smoothness criterion}, we compute: 
    \begin{align}
        N^{-1}\|DPf_k(X)\|_{F}^{2} &= N^{-1}\sum_{i=1}^{N}\|DPf_k(x_i)\|_{2}^{2} \\
        &= {\textsf Tr}\left(P^{\top}D^{\top}DP\left[N^{-1}\sum_{i=1}^{N}f_k(x_i)f_k(x_i)^{\top}\right]\right) \\
        &=: {\textsf Tr}\left(P^{\top}D^{\top}DP\Sigma_{N}\right). \label{eq: descrambling function with sample autocorrelation}
    \end{align}
    Here $\Sigma_N$ is the empirical autocorrelation matrix given by 
    \begin{align}
        \Sigma_{N} := N^{-1}\sum_{i=1}^{N}f_k(x_i)f_k(x_i)^{\top}. \label{eq: autocorrelation matrix}
    \end{align}
    Note that $\Sigma_N \to \Sigma := \mathbb{E}_{x}[f_{k}(x)f_k(x)^{\top}]$ almost surely as $N \to \infty$ due to the law of large numbers. Next we define $\delta = \inf\{\lambda_{i+1} - \lambda_{i}\}$ where $\lambda_i$ are the eigenvalues of $\Sigma$. Since by assumption these are distinct, $\delta > 0$. Furthermore, since the eigenvalues of $\Sigma_N$ converge to those of $\Sigma$ (see, for instance \cite[Theorem 6, Chapter 9]{lax07}) the eigenvalues of $\Sigma_N$ are also distinct for large enough $N$. Assuming such an $N$ large enough, the eigenspaces of $\Sigma_{N}$ are all distinct. Then by the modified Davis-Kahan theorem \cite[Theorem 2]{10.1093/biomet/asv008}, for each index $j$, there exists an eigenvector of $u^{j}_{N}$ of $\Sigma_N$ corresponding to the $j$th eigenspace such that 
    \begin{align}
        \|u^{j}_{N} - u^{j}\|^{2}_{2} \leq \frac{8\|\Sigma_{N} - \Sigma\|^{2}_{F}}{\delta^2}.
    \end{align}
    Here $u^j$ is the $j$th eigenvector of $\Sigma$. Now iterating the above bound over all the $n_k$ eigenvectors $u^{j}_{N}$'s we get that there exist eigenvectors of $U_N$ of $\Sigma_N$ such that 
    \begin{align}
        \|U_N - U\|_{F} \leq \frac{2^{3/2}\sqrt{n_k}\|\Sigma - \Sigma_N\|_{F}}{\delta}.  \label{eq: perturbation bound}
    \end{align}
    Now, we may write $\widehat{P}_N = TU^{\top}_N$ and from Lemma \ref{lem:brockett} verify that $\widehat{P}_N$ minimizes the objective function \eqref{eq: descrambling function with sample autocorrelation}. As a consequence we have that $\widehat{P}_N \to TU^{\top}$ almost surely. 
 \end{proof}

\subsection{Variations in input distribution and network architecture}

 We apply Theorem \ref{thm: main theorem} to a variety of conditions for $f_k$ and distributions of the input-output pairs $(x,y)$. We additionally define the hidden layer function $h_k(x)$ by  
 \begin{align}
     h_k(x) = \phi(f_{k-1}(x)). \label{eq: hidden data}
 \end{align}
 Here $\phi$ is the nonlinear activation function. To obtain the hidden layer data $h_k(X)$ we apply the function $h_k$ column-wise $X$. Note that $f_k(X) = W_k h_k(X)$. Then since the columns of $h_k(X)$ are i.i.d. samples of $h_k(x)$, the distribution of $h_k(x)$ can be prescribed and the limiting behaviour of the descramblers $\widehat{P}(k,X,N)$ can be studied. We start with the case where hidden data the $h_k(x)$ has a whitened distribution with autocorrelation matrix $I$. Whitened hidden data is a common assumption in many theoretical analyses of NNs \cite{braun2022exact, ghorbani2019limitations}. To apply Theorem \ref{thm: main theorem} we must understand the eigendecomposition of $D^{\top}D$ where $D$ is a second-order finite-difference/Fourier differentiation matrix of dimension $n_k$. In both cases, $D$ is a real and circulant matrix so it will admit singular vectors given by the Hartley transform \cite{karner2003spectral}. For example, if $D$ is a periodic finite-difference Laplacian then the columns of $T$ sorted in ascending order of eigenvalues can be characterized as follows: 
 \begin{align}
    \label{eq: real fourier matrix}
        T_{m}(l) = \begin{cases} \frac{1}{\sqrt{n_k}}\left(\cos{\frac{\pi ml}{n_k}} + \sin{\frac{\pi ml}{n_k}}\right), \: \: \quad m \text{ even, } \\
        \frac{1}{\sqrt{n_k}}\left(\cos{\frac{\pi (m+1)l}{n_k}} + \sin{\frac{\pi (m+1)l}{n_k}}\right), \: \: \quad m \text{ odd}.\end{cases} 
\end{align}

The following result shows that in the large data limit, the action of the descrambled weight matrix is given by basis projection followed by expansion in the trigonometric basis given by \eqref{eq: real fourier matrix}:

\begin{theorem}[Isotropic data] 
\label{thm: isotropic data}
Let $f_k$ be given by \eqref{eq: wiretap}. If $n_{k-1} \geq n_k$ and the singular values of $W_k$ are distinct, then there exist descramblers $\widehat{P}_{SC}(k,X,N)$ such that the descrabled weight matrices $\widehat{P}_{SC}(k,X,N)W_k \to \widehat{\mathcal{W}}_k$ as $N \to \infty$ where the action of $\widehat{\mathcal{W}}_k$ is given by: 
\begin{align}
    \widehat{\mathcal{W}}_k x = \sum_{n=1}^{n_k}T_n \sigma_n \langle v_n, x \rangle \label{eq: linear network, noise only},
\end{align}
where $v_n$ is the right singular vector of $W_k$ and $\sigma_n$ is the corresponding singular value. 
\end{theorem}

 \begin{proof}
     The sample autocorrelation matrix $\Sigma_N$ of the data $f_k(X)$ is given by
     \begin{align}
         \Sigma_N := W_k\left[\frac{1}{N}\sum_{i=1}^{N}(h_k(x))_{i}(h_k(x))_{i}^{\top}\right]W_{k}^{\top}.
     \end{align}
     Now, since $h_{k}(x_i)$ are samples of a zero mean isotropic random variable, we get that $\Sigma_N \to W_kW^{\top}_k$ almost surely. Since $W_k$ has distinct singular values and its output dimension $n_k \leq n_{k-1}$ by assumption, we get that the eigenvalues of $W_kW^{\top}_k$ are distinct. Therefore, Theorem \eqref{thm: main theorem} applies and we have that there exists a sequence of descramblers $\widehat{P}_{SC}(1,X,N)$ such that $\widehat{P}_{SC}(1,X,N)W_1 \to TU^{\top}W_1$ where $U$ is the matrix of singular vectors of $W_1$. Relabelling $TU^{\top}W_1$ as $\widehat{\mathcal{W}}_1$ we have that $\widehat{\mathcal{W}}_1 = TU^{\top}U \Sigma V^{\top} = T\Sigma V^{\top}$ where $W_1 = U\Sigma V^{\top}$ is the SVD of $W_1$. 
 \end{proof}
\begin{remark}
    Theorem \ref{thm: isotropic data} shows that NN descrambling mathematically captures the process of representing the directions of largest variance of the preactivation data through trigonometric functions of lowest energy.  
\end{remark}
\begin{remark}
     Although the descramblers $\widehat{P}_{SC}(k,X,N)$ may be non-unique, this degeneracy may be circumvented in the large data limit in the sense that the action of the descrambled weights can be described by projection onto one-dimensional subspaces followed by expansion in a trigonometric basis. We must also assume the distinctness of the singular values of the raw/scrambled weights $W_k$. In practice this is not a restriction since neural networks whose weights have distinct singular values are dense in the space of all networks.  
\end{remark}
Next we consider the case when the hidden layer vectors $h_k$ are noisy measurements of some parameter (with a given prior) under isotropic noise with a prescribed SNR $\alpha$. Specifically, let $z \in \mathbb{R}^d$ be a random variable with finite moments. Let $s: \mathbb{R}^d \to \mathbb{R}^{n_k}$ such that $\mathbb{E}_{z}[\|s(z)\|_{2}^{2}] < \infty$. We assume that distribution of hidden layer inputs is given by $h_k(x) \sim s(z) + \alpha^{-1} \xi$. This situation arises widely in inverse problems and signal processing, and indeed the original motivation for NN descrambling was to study a network trained to recover $z$ from $x = h_0$. In this case, we show that as the data gets noisier, the descrambled weight matrices converge to the form given in \eqref{eq: linear network, noise only}. 

\begin{theorem}[Noisy signal input] \label{thm: noisy signal input} Let $f_k$ be given by \eqref{eq: wiretap} and assume $W_k$ has distinct singular values where $n_k \leq n_{k-1}$. Let $h_k(x) \sim s(z) + \alpha^{-1} \xi$ where $z$ and $\xi$ are independent distributions, $\xi$ is isotropic and zero mean. Then for small enough $\alpha$, as $N \to \infty$ there exist descrambled weights $\widehat{P}_{SC}(k,X,N)W_k \to \widehat{\mathcal{W}}_k$ where for every $x \in \mathbb{R}^{n_k}$,   
\begin{align}
    \widehat{\mathcal{W}}_k x = \sum_{i=1}^{n_k}T_n \sigma_n \langle v_n, x\rangle + O(\alpha^{2}). \label{eq: signal + noise}
\end{align}
\begin{proof}
    Note that multiplying the hidden data by $\alpha$ will not change the minimizers in \eqref{eq: descrambling equation}. Due to the zero mean property and isotropy of $\xi$ we get that $\mathbb{E}[h_k h_k^{\top}] = \mathbb{E}[s(z)s(z)^{\top}] + \alpha^{-2}I$. Therefore the autocorrelation matrix of the rescaled hidden data $\alpha f_k = \alpha W_k h_k$ is $\Sigma_{\alpha} = \alpha^2 W\mathbb{E}[s(z)s(z)^{\top}]W^{\top} + WW^{\top}$. For small enough $\alpha$, $\Sigma_{\alpha}$ will have distinct eigenvalues because it is a small perturbation of $WW^{\top}$. Therefore, Theorem \ref{thm: main theorem} applies and we get that the descramblers $\widehat{P}(k,X,N) \to TU^{\top}_{\alpha}$ where $U_{\alpha}$ are the eigenvectors of $\Sigma_{\alpha}$. But since $\Sigma_{\alpha} = WW^{\top} + O(\alpha^{2})$ we can use the Davis Kahan bound \eqref{eq: perturbation bound} on $WW^{\top}$   and $\Sigma_{\alpha}$ to conclude that $\|U_{\alpha} - U\| = O(\alpha^{2})$. Since the large data limit descrambled weights are given by $TU^{\top}_{\alpha}W_1$, \eqref{eq: signal + noise} holds readily.  
\end{proof}
\end{theorem}

\subsection{NN descrambling in the presence of training}

Finally, we tackle a case where we unite NN descrambling with NN training. Here we make a number of simplifying assumptions: 
\smallskip

    \noindent (A1) The NN is a \emph{two-layer linear network} with one hidden layer, i.e the NN is $f = W_2 W_1 x$ and the $f_k$ under consideration for descrambling are $f_1 = W_1 x$ and $f_2 = f$. Such two-layer linear networks are examples of the highly-studied deep linear networks \cite{saxe2014exact, saxe2019mathematical, braun2022exact}. 
    
    \noindent (A2) $f$ is trained with finitely many examples $(x_i, y_i)_{i=1}^{N}$ with mean squared error loss: 
    \begin{align}
        \mathcal{L}(W_1,W_2) = \frac{1}{N}\sum_{i=1}^{N}\|y_i - W_2 W_1 x_i\|_{2}^{2}.
    \end{align}
    The loss $\mathcal{L}$ is minimized under a \emph{gradient flow} with learning rate $1/\tau$: 
    \begin{align}
        \tau \frac{dW_1}{dt} = -\nabla_{W_1}\mathcal{L}, \: \tau \frac{dW_2}{dt} = -\nabla_{W_2}\mathcal{L}. \label{eq: gradient flow}
    \end{align}
    
    \noindent (A3) The eigenvectors of the input-input correlations $\Sigma_{N}^{xx}$ coincide with the right singular vectors of the input-output correlations $\Sigma_{N}^{xy}$:
    \begin{align}
        \Sigma_{N}^{xx} = \frac{1}{N}\sum_{i=1}^{N}x_{i}x^{\top}_{i}, &\quad \Sigma_{N}^{xy} = \frac{1}{N}\sum_{i=1}^{N}y_{i}x^{\top}_{i}, \\
        \Sigma_{N}^{xx} = V_{N}\Lambda_{x}V_{N}^{\top}, &\quad \Sigma_{N}^{xy} = U_{N}\Lambda_{xy}V_{N}^{\top}.
    \end{align}
    
    \noindent (A4) The weights $W_1, W_2$ are initialized with an initial condition 
    \begin{align}
        W_{1} = RD_1V^{\top}_{N}, \quad W_{2} = U_{N}D_2R^{\top}. \label{eq: initial condition}
    \end{align}
    Here $R$ is an arbitrary orthogonal matrix.
    In \cite{saxe2019mathematical} and \cite{saxe2014exact} it is shown that the NN weights initialized using layerwise pretraining or random orthogonal initialization approximately satisfy the conditions \eqref{eq: initial condition} when trained under the gradient flow \eqref{eq: gradient flow}. Here we assume that the matrices $W_1$ and $W_2$ satisfy this form exactly to simplify our calculations.   

The assumptions $(A1)-(A4)$ provide a setting in which the dynamics of two-layer linear networks can be described exactly. 

\begin{proposition}[\cite{saxe2014exact}]
\label{prop: saxe}
    Under the assumptions $(A1-A4)$ the dynamics of $W_1(t)$ and $W_2(t)$ can be described by the following formula: 
    \begin{align}
        W_1(t) = R\Lambda_{1}(t)V^{\top}_{N}, \quad W_2(t) = U_{N}\Lambda_{2}(t)R^{\top}. \label{eq: saxe}
    \end{align}
    Here $R$ is the arbitrary fixed orthogonal matrix used for initialization in \eqref{eq: initial condition} and $\Lambda_{1}, \Lambda_{2}$ are diagonal matrix-valued functions defined on $[0,+\infty)$. 
\end{proposition}

\begin{theorem}
\label{thm: NN descrambling with training}
    Suppose $(A1-A4)$ are satisfied. Then for every $t \in [0, +\infty)$ there exist descrambled weight matrices given by: 
    \begin{align}
        \widehat{P}(1,X,N)W_1 = T\Lambda_1V^{\top}_{N}, \quad \widehat{P}(2,X,N)W_2 = T\Lambda_2R^{\top}. \label{eq: SGM descrambling}
    \end{align}
\end{theorem}

\begin{proof} (Theorem \ref{thm: NN descrambling with training}) 
The NN weights $W_2$ and $W_1$ are described by \eqref{eq: saxe}. Denoting by $X$ the $n_0 \times N$ matrix of i.i.d samples of $x$, we proceed layer by layer. For $k=1$, let $\Sigma_N$ be the autocorrelation matrix of $S = W_1X$. Note that due \eqref{eq: saxe}, $\Sigma_{N} = R\Lambda_{1}(t)R^{\top}$. Then from the proof of Theorem \ref{thm: main theorem} we have that for fixed $N$ the descramblers are given by $TR^{\top}$ and the descrambled weights are described by $TR^{\top}W_1 = T\Lambda_{1}V^{\top}_{N}$. For $k=2$, the autocorrelation matrix of the finite samples is given by $U_{N}\Lambda(t)U_{N}^{\top}$ so the descramblers are given by $TU^{\top}_{N}$ and the descrambled weights by $T\Lambda_{2}R^{\top}$. 
\end{proof}

In Theorem \ref{thm: NN descrambling with training} if the sample input-input correlation matrix converges to a limit with distinct eigenvalues then Theorem \ref{thm: main theorem} may be applied to obtain a limiting descrambled matrix for the first layer. Theorem \ref{thm: NN descrambling with training} shows that under the SMG model, the descrambled first layer reflects the data (through the influence of $V^{\top}_N$) while the descrambled second layer reflects the influence of the network \emph{initialization} (through the influence of $R$). The assumptions A1, A2, and A4 are regularly used settings for studying the training dynamics of neural networks. The assumption A3 on the training data is quite restrictive and is usually not satisfied exactly. However, it has been illustrated numerically \cite{saxe2019mathematical,baldi1989neural} that the training dynamics \eqref{eq: saxe} are stable with respect to small deviations in A3. This means that the NN still approximately obeys \eqref{eq: saxe} when A3 is approximately satisfied. A case of approximate satisfaction of A3 is given by noisy linear measurements of a signal under isotropic noise. Let the random variable $z \in \mathbb{R}^{n_2}$ model a signal to be estimated such that $\mathbb{E}[zz^{\top}] = kI$ for some $k \in \mathbb{R}$. Moreover, let  $A: \mathbb{R}^{n_2} \to \mathbb{R}^{n_0}$ be a linear transformation and $\xi$ a zero-mean isotropic random variable. Set the NN inputs to be i.i.d samples from $x = Az + \xi$ and the outputs as $y = z$. Then the training data $(x,y)$ satisfies assumption $A3$: 
\begin{align}
    \Sigma^{yx} = \mathbb{E}[zz^{\top}]A^{\top} = kA^{\top}, \\
    \Sigma^{xx} = A\mathbb{E}[zz^{\top}]A^{\top} + I = kAA^{\top} + I. \label{eq: approximate autocorrelation} 
\end{align}
$[\Sigma^{yx}]^{\top}\Sigma^{yx}$ and $\Sigma^{xx}$ have the same eigenspaces. As a consequence, $N$ i.i.d samples of $(x,y)$ will approximately satisfy the condition A3. This observation suggests that descrambling NNs trained for noisy signal estimation (such as DEERNet) might reveal the training dynamics of the weights. We will provide a numerical illustration of this idea in Section \ref{subsec: training dynamics}. 

\subsection{Descrambling as rearrangement of neurons}

The action of the descrambler $\widehat{P}_{SC}(k,X,N)$ on the weights $W_k$ can be modeled as a symmetry that ``rearranges" the neurons in the $k$th layer for optimal explainability. We describe a case where this rearrangement is a permutation. Notably, this occurs when using convolutional networks. 

\begin{proposition}[Convolutional NNs] \label{prop: CNNs}
Let $f_k$ be given by \eqref{eq: wiretap} and assume the hidden data $h_k(X)$ is whitened. Moreover, assume $W_k$ represents a convolution with stride 1 where $W_k x = w * x$. Then $\widehat{P}(k,X,N)W_1 = \Pi^{\top} W_1$ where $\Pi^{\top}$ is a permutation. 
\end{proposition}

\begin{proof}
    Let $\Sigma^{h}_{N}$ be the autocorrelation matrix of the data $h_k(X)$. By assumption $\Sigma^{h}_{N} = I$ so the autocorrelation matrix of the preactivation data $f_k(X)= W_kh_k(X)$ is $\Sigma_N = W_k W_{k}^{\top}$ \eqref{eq: autocorrelation matrix}. Since $W_k$ represents a circular convolution $W_kW^{\top}_{k}$ must be diagonalized by the real Fourier basis given by $T$ in \eqref{eq: real fourier matrix} \cite{karner2003spectral}. To apply Theorem \ref{thm: main theorem} we must rearrange the columns of $T$ through a permutation matrix $\Pi$ to match the descending order of the eigenvalues of $W_kW_k^{\top}$. Representing this rearranged eigenbasis as $T\Pi$ the descramblers are given as $\widehat{P}(k,X,N) = T(T\Pi)^{\top} = T\Pi^{\top}T^{\top} = \Pi^{\top}$ since $TT^{\top} = I$. 
\end{proof}

 \begin{remark}
     The assumption of whitened finite samples of hidden data $h_{k}(X)$ is stronger than the assumption of whitened distributions of hidden data made in Theorem \ref{thm: isotropic data}. However, there are several contexts in which hidden data is isotropic or made to be isotropic for enhancing the learning capabilities of fully connected NNs \cite{luo2017learning,wiesler2011convergence, ye2019network}. 
 \end{remark}

\section{Numerical results}

In this section we will descramble the network \textsf{DEERNet} given by the architecture $\textsf{DEERNet}(x) = \textsf{logsigmoid}(W_2\textsf{tanh}(W_1x))$ as described in \cite{amey2021neural} \footnote{The code to generate data and figures has been published at \url{https://github.com/ShashankSule/ESVD}}. The NN was trained to solve the following integral equation arising in double electron-electron resonance (DEER):
\begin{align}
        \Gamma(t) = \int_{\Omega}p(r)\gamma(r,t)\,dr + \xi. \label{eq: deer equation}
\end{align}
Here $\xi$ is considered to be white noise with a preset signal-to-noise ratio (SNR) and $\gamma(r,t)$ is the DEER kernel given by
{\normalsize
\begin{equation}
\label{eq: deer model}
    \gamma(r,t)\! := \!\sqrt{\frac{\pi}{6At}}\left[\cos[At]C\!\!\left[\sqrt{\frac{6At}{\pi }}\right] \!+\! \sin[At]S\!\!\left[\sqrt{\frac{6At}{\pi }}\right]\right],
\end{equation}
\begin{equation}
    A := \frac{\mu_0}{4\pi}\frac{\gamma_1 \gamma_2 h}{r^3}; \: C(x) = \int_{0}^{x}\cos(t^2)\,dt; \: S(x) = \int_{0}^{x}\sin(t^2)\,dt .
\end{equation}}%
\textsf{DEERNet} was trained to approximate the solution operator taking $\Gamma$ to $p$ using synthetically generated input-output pairs $(\Gamma_{i}, p_i)$. 

\subsection{Non-uniqueness of descramblers}

In order to apply descrambling to a specific NN we first have to address the issue of non-uniqueness of descramblers. Lemma \ref{lem:brockett} combined with \eqref{eq: brockett to descrambling} shows that the solutions to the NN descrambling problem are only unique up to the choices of orthogonal bases for the singular subspaces of $SS^{\top}$ (the data autocorrelation matrix) and $D^{\top}D$ (the differentiation stencil, or more generally, a context dependent linear operator). A scheme to select a particular minimizer to act as an explanation for the weights $W_k$ was formulated in \cite{amey2021neural} where it was proposed that the smoothness objective function $\eta_{SC}$ be minimized over $O(n)$ through the Cayley transform: 
\begin{align}
P(Q) = (I - Q)(I + Q)^{-1}, \: Q = -Q^{\top}. \label{eq: cayley transform}
\end{align}
The Cayley transform is an injective parameterization of the Lie group $O(d)$ through its Lie Algebra $S^{-}_{d}(\mathbb{R})$, the vector space of skew-symmetric matrices on $\mathbb{R}$. Through the Cayley transform, the smoothness descrambling problem on the orthogonal group may be restated as a nonlinear optimization problem on the vector space $S^{-}_{d}(\mathbb{R})$:
\begin{align}
    \widehat{Q}(k,X,N) = \underset{Q \in S^{-}_{d}(\mathbb{R})}{\textsf{argmin}} \: \|D(I-Q)(I+Q)^{-1}f_k(X)\|_{F}^{2} := \underset{Q \in S^{-}_{d}(\mathbb{R})}{\textsf{argmin}} G(Q). \label{eq: cayley reparameterization}
\end{align}
The advantage of the formulation \eqref{eq: cayley reparameterization} is that now the descramblers $\widehat{P}(k,X,N)$ may be found using a gradient flow:
\begin{align}
    Q'(t) &= \tau \nabla_{S^{-}_{d}}G(Q(t)), \quad Q(0) = Q_0, \label{eq: Cayley gradient flow}\\
    \widehat{Q}(k,X,N) &= \widehat{Q}_{Q_0} := \lim_{t \to \infty} Q(t).
\end{align}

\begin{figure}[h]
    \centering
\includegraphics[width=0.8\textwidth]{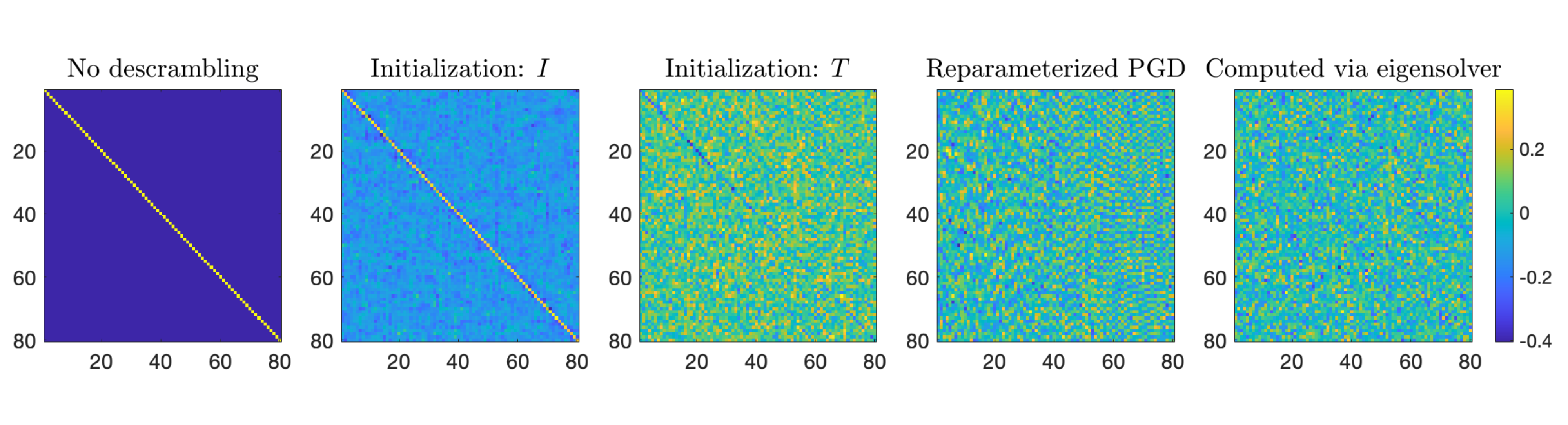}
    \caption{Left to Right: The descamblers $\widehat{P}_{SC}(1,X,N)$ were computed using four different strategies (a) Projected Gradient Descent (PGD) on \eqref{eq: Cayley gradient flow} with $P(Q_0) = I$, (b) Warm start, i.e PGD with $P(Q_0) = T$, (c) Rescaled PGD where $D^{\top}D = T\Omega T^{\top}$ was replaced with a diagonal matrix $\Omega$ and solutions were interpreted through multiplication by $T$, and (d) Direct eigendecomposition of $S = f_1(X) = W_1X$.}
    \label{fig: descramblers_5panel}
\end{figure}
\begin{figure}[h]
    \centering
    \includegraphics[width=0.8\textwidth]{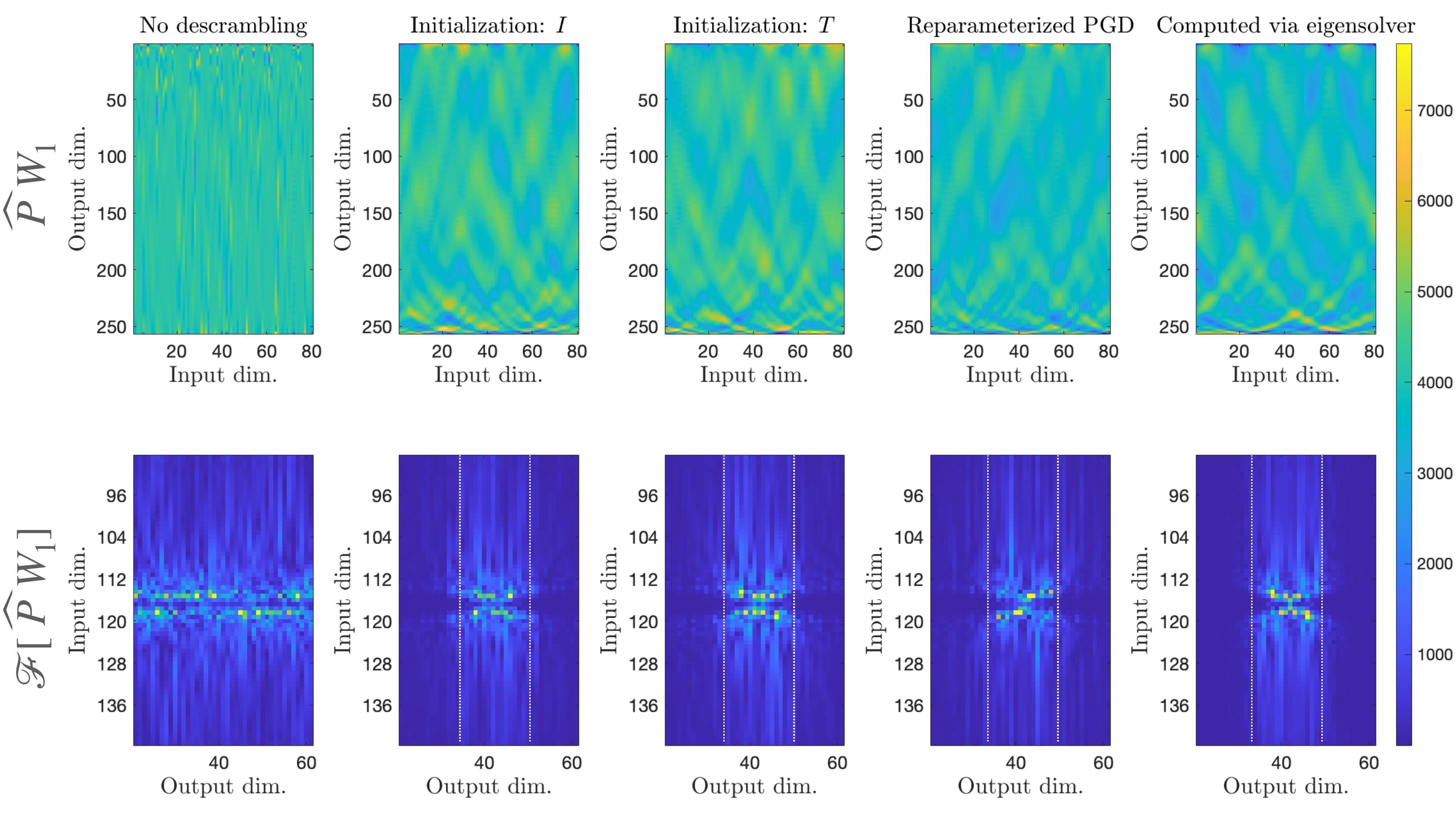}
    \caption{Top: Descrambled weight matrices $\widehat{P}_{SC}(1,X,N)$ computed for the four strategies outlined in Figure \ref{fig: descramblers_5panel}. Notably, while the descramblers are quite different, the descrambled weights themselves are quite similar. Bottom: This similarity among explanations can be quantified by moving to the Fourier domain where nearly all descrambled weights have a notch at the zero frequency and a bandpass filter along the output dimension in approximately the $\omega_1 \in [-10,10]$ frequency band.}
    \label{fig: 5panel_eigensolver}
\end{figure}

 Given that $G(Q)$ has multiple minimizers, the choice of the initial point $Q_0$ is essential for determining the solution $\widehat{Q}$ as long as the stepsize $\tau$ is sufficiently small. Alternately, following Theorem \ref{thm: main theorem} the gradient-based minimization approach \eqref{eq: Cayley gradient flow} may be discarded entirely in favour of using an eigensolver, computing $U_S$ and then writing the descrambler as $\widehat{P} = TU^{\top}_{S}$ for some fixed choice of $T$. In Figure \ref{fig: descramblers_5panel} we compare these approaches to finding the descramblers of the first layer of \textsf{DEERNet} by exploring three different initializations for the gradient flow \eqref{eq: Cayley gradient flow} and directly using an eigensolver for descrambling. We find that although the descamblers are different in all four cases, the explanations of $W_1$ they produce upon multiplication are nearly identical. In Figure \ref{fig: 5panel_eigensolver} it can be observed that the time-domain representations of the descrambled matrices $\widehat{P}(1,X,N)W_1$ feature wave-like interlocking patterns. To understand these patterns we use the trick in \cite{amey2021neural} by inspecting the Fourier domain representations of these matrices as filters, revealing a notch filter at the base frequency. However, we find that the different methods of computing the descramblers lead to slightly different qualities of notch filter. In particular, the weight matrix descrambled through the eigensolver has a sharp frequency cutoff outside the $[-10,10]$ band. Although the weight matrices descrambled through the three other methods also contain the same bandpass filtering feature, the frequency cutoff is certainly noisier and less pronounced than the eigensolver method (Table \ref{tab: bandpass filter}). As a consequence, directly computing the solutions to \eqref{eq: descrambling equation} via Theorem \ref{thm: main theorem} may be more beneficial for explainability than using the gradient-descent based method suggested in \cite{amey2021neural}. 
 


\begin{table}[h]
\begin{tabular}{llllll}
\hline
& No desc.  & Init.: $I$ & Init.: $T$ & Reparam   & Eigensolver \\ \hline
\begin{tabular}[c]{@{}l@{}}Rel. norm of \\ coefficients  where \\ $\omega_1 \notin [-10,10]$\end{tabular} & 7.86$ \times 10^{-1}$ & 5.87$ \times 10^{-2}$  & 5.40$ \times 10^{e-2}$  & 7.63$ \times 10^{-2}$ & \textbf{9.66$ \times 10^{-3}$} \\ \hline
\end{tabular}
\vspace{0.1in}
\caption{Figure \ref{fig: 5panel_eigensolver} displays a bandpass in the $\omega_1 \in [-10,10]$ frequency band in the descrambled weights. We additionally find that in the descrambled weights, less than 5\% of the norm is concentrated outside this band. Moreover, the weights descrambled via the eigensolver have the sharpest band cut off. \label{tab: bandpass filter}}
\end{table}

\subsection{Assessing the validity of interpretation}
\label{subsec: training dynamics}
\begin{figure}
    \centering
    \includegraphics[width=0.8\textwidth]{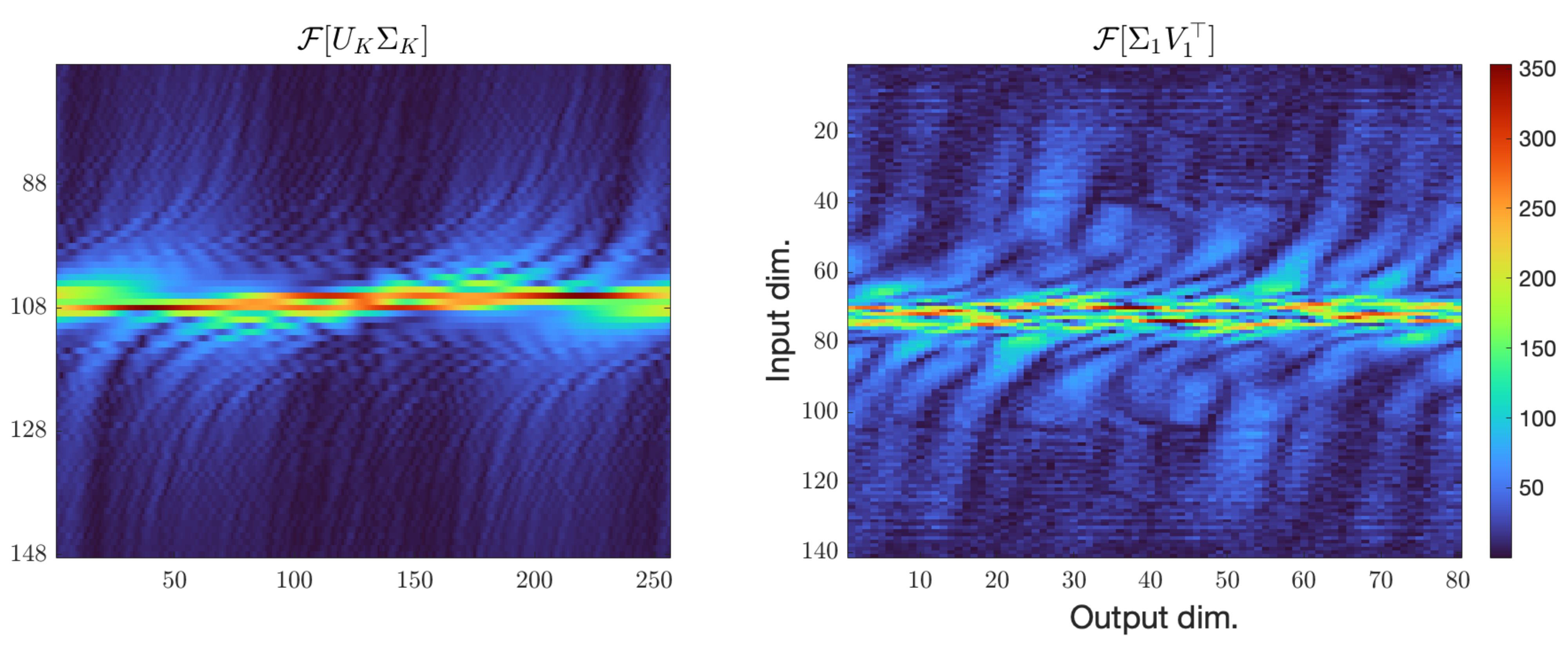}
    \caption{Left: The 2D FFT of the singular vectors of the discretized integral kernel in \eqref{eq: deer equation}. Right: The 2D FFT of the scaled right singular vectors of the trained weights. Note that the panel on the right resembles a noisier version of the left panel, suggesting that the first layer of the NN acts as a pseudo-inverse.}
    \label{fig: 2panel}
\end{figure}

To what extent is the notch filter in Figure \ref{fig: 5panel_eigensolver} inherent to the network weights $W_1$ and not an artefact of (a) the influence of the differentiation matrix $D$, or (b) the training data $X$? We explore this question by using Theorem \ref{thm: NN descrambling with training}. In particular, suppose DEERNet was a two-layer linear network given by $f = W_2W_1x$. Since it is trained on data of the form $x = Kz + \alpha^{-1}\xi$ where $K$ is the discretization of an integral kernel, if $\mathbb{E}[zz^{\top}] = I$ then the solutions to NN descrambling given by Theorem \ref{thm: NN descrambling with training} would approximately hold and as a result, the descrambled weight matrix $\widehat{P}W_1$ should be well approximated by the matrix $T\Sigma_{W}V^{\top}_{N}$ where $V^{\top}_{N}$ are the eigenvectors vectors of the input autocorrelation matrix. This autocorrelation matrix would in turn be well-approximated by $KK^{\top}$ (see \eqref{eq: approximate autocorrelation}). Thus, we should have that $V_N \approx U_K$, where $U_K$ are the eigenvectors of $KK^{\top}$. In Figure \ref{fig: 2panel} we confirm that this Ansatz is good: in particular, we find cubic streaks in the Fourier domain representation of $\Sigma_1 V^{\top}_{1}$ mimicking to those in the Fourier domain representation of $U_{K}\Sigma_K$, thus validating the interpretation of (a part of) $W_1$ as a notch filter in Figure \ref{fig: 2panel}. Thus, the projection induced by principal components--now viewed as descrambling--helps us identify a model for the training process of the first layer.  

\subsection{Discovering motifs within singular vectors of higher layers}

\begin{figure}[h]
    \centering
    \includegraphics[width=0.8\textwidth]{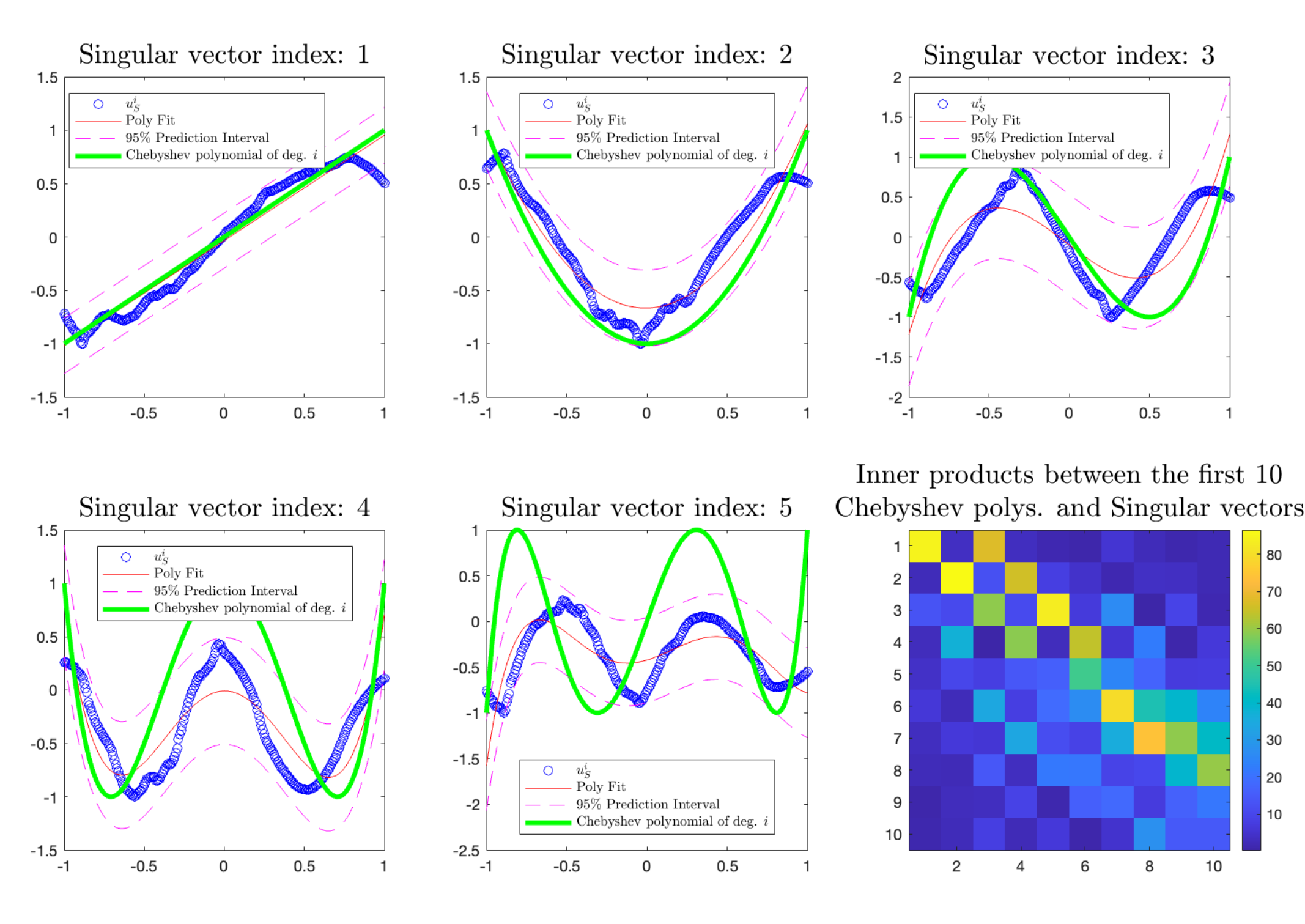}
    \caption{The top five principal components $\{u^{i}_{S}\}_{i=1}^{5}$ of the second layer preactivation data $S = W_2 f_1(x)$ resemble a system of orthogonal polynomials. In \cite{amey2021neural} a powerful plausibility argument suggested these can be fit by Chebyshev polynomials. Bottom right: After shifting and rescaling the principal components, the inner product matrix between the top 20 Chebyshev polynomials $C_n$ and the principal components $u^{i}_{S}$ is approximately banded.}
    \label{fig: 6panel_chebyshev}
\end{figure}

Theorems \ref{thm: isotropic data} and \ref{thm: noisy signal input} are limited due to the restrictive assumptions on the distribution of the hidden layer activations. Usually these assumptions are only satisfied by the training data itself, thus making the theoretical characterizations of the descramblers $\widehat{P}_{SC}(k,X,N)$ only applicable for the first layer. Nonetheless, in \cite{amey2021neural} it was illustrated that the \emph{second} layer of DEERNet regularizes the incoming signal by projection onto the Chebyshev basis. However, these features were discovered by descrambling both the rows and columns of the second weight matrix $W_2$ simultaneously. Here we illustrate that the emergence of this orthogonal basis is much more straightforward: in particular, by resorting to Theorem \ref{thm: main theorem} we visualize the left singular vectors $U_S$ of the second layer activations $S = f_2(X)$. We discover that the largest singular vectors of $U_S$ resemble a system of orthogonal polynomials. This is because (1) the singular vectors at odd (resp. even) indices tend to resemble odd (resp. even) functions and (2) the singular vectors, although oscillating, are off-phase so a polynomial structure is more plausible than a trigonometric structure. We confirm this fact in Figure \ref{fig: 6panel_chebyshev}. We also test the hypothesis that these singular vectors represent noisy realizations of the Chebyshev polynomials $C_i$ by computing the pairwise inner products $\langle C_i, u^{j}_{S} \rangle$ and find a banded structure in the inner product matrix, suggesting that the $j$th singular vector may indeed reside in the subspace formed by some $C_{j-k},\ldots, C_{j+k}$ for some small $k$. In any case, here we have illustrated that a complicated optimization procedure via the Cayley transform may be circumvented in favour of an eigendecomposition. This eigendecomposition, when viewed by a subject matter expert, can nevertheless lead to helpful explanations of the mechanisms of the NN at the $k$th layer. 

\section{Conclusion}


We have illustrated that the SVD plays a key role in explaining NN weights via NN descrambling. Moreover, the explanations so produced depend considerably on the training dynamics, input data, network architecture, and even the initialization. We have shown that the empirical results from \cite{amey2021neural} can be recreated using eigendecompositions, showing that the principal components of hidden data can be used for visually explaining and reverse-engineering the hidden actions of individual layers. However, the emergence of the SVD reveals a key drawback of this method in the cases where the required interpretation may be nonlinear or where mathematical patterns may be present across layers rather than within layers. Fixing these drawbacks necessitates extensions of NN descrambling to alternative types of explainability loss functions and less elegant group structures. These avenues are highly promising and are the subject of future work. 



\begin{thebibliography}{555}
\bibitem{yosinski2015understanding}
Yosinski, J., Clune, J., Nguyen, A., Fuchs, T. \& Lipson, H. Understanding neural networks through deep visualization. (2015)
\bibitem{simonyan2014deep}
Simonyan, K., Vedaldi, A. \& Zisserman, A. Deep inside convolutional networks: visualising image classification models and saliency maps. 
\bibitem{nanda2023progress}
A, N., Chan, L., Lieberum, T., Smith, J. \& Steinhardt, J. Progress measures for grokking via mechanistic interpretability. {\em Arxiv Preprint Arxiv:2301.05217}.(2023)
\bibitem{kadkhodaie2023generalization}
Kadkhodaie, Z., Guth, F., Simoncelli, E. \& Mallat, S. Generalization in diffusion models arises from geometry-adaptive harmonic representation. 
{\em Arxiv Preprint Arxiv:2310.02557}.(2023)
\bibitem{amey2021neural}
Amey, J., Keeley, J., Choudhury, T. \& Kuprov, I. Neural network interpretation using descrambler groups. 
{\em Proceedings Of The National Academy Of Sciences}. \textbf{118} (2021)
\bibitem{trefethen2000spectral}
Trefethen, L. Spectral methods in MATLAB. (SIA,2000)
\bibitem{worswick2018deep}
Worswick, S., Spencer, J., Jeschke, G. \& Kuprov, I. Deep neural network processing of DEER data. {\em Science Advances}. \textbf{4}, eaat5218 (2018)
\bibitem{brockett1989least}
Brockett, R. Least squares matching problems. {\em Linear Algebra And Its Applications}.\textbf{122} pp. 761--777 (1989)
\bibitem{absil2008optimization}
Absil, P., Mahony, R. \& Sepulchre, R. Optimization algorithms on matrix manifolds. (Princeton University Pres,2008)
\bibitem{lax07}
Lax, P. Linear Algebra and Its Applications. (Wiley-Interscience,year = 20)
\bibitem{10.1093/biomet/asv008}
Yu, Y., Wang, T. \& Samworth, R. A useful variant of the Davis–Kahan theorem for statisticians. {\em Biometrika}. \textbf{102}, 315-323 (2014)
\bibitem{braun2022exact}
Braun, L., Domin{\'e}, C., Fitzgerald, J. \& Saxe, A. Exact learning dynamics of deep linear networks with prior knowledge. {\em Advances In Neural Information Processing Systems}.\textbf{35} pp. 6615--6629 (202)\
\bibitem{ghorbani2019limitations}
Ghorbani, B., Mei, S., Misiakiewicz, T. \& Montanari, A. Limitations of lazy training of two-layers neural network. {\em Advances In Neural Information Processing Systems}. \textbf{32} (201)
\bibitem{karner2003spectral}
Karner, H., Schneid, J. \& Ueberhuber, C. Spectral decomposition of real circulant matrices. {\em Linear Algebra And Its Applications}. \textbf{367} pp. 301--311 (2003)
\bibitem{saxe2019mathematical}
Saxe, A., , J. \& Ganguli, S. A mathematical theory of semantic development in deep neural networks.{\em Proceedings Of The National Academy Of Sciences}. \textbf{116}, 11537--11546 (2019)
\bibitem{baldi1989neural}
Baldi, P. \& Hornik, K. Neural networks and principal component analysis: Learning from examples without local minima. {\em Neural Networks}. \textbf{2}, 53--58 (1989)
\bibitem{luo2017learning}
Luo, P. Learning deep architectures via generalized whitened neural networks. In: International Conference on Machine Learning. PMLR. 2017, pp. 2238–2246.
\bibitem{wiesler2011convergence}
Wiesler, S. \& Ney, H. A convergence analysis of log-linear training. 
{\em Advances In Neural Information Processing Systems}. \textbf{24} (201)
\bibitem{ye2019network}
Ye, C., Evanusa, M., He, H., Mitrokhin, A., Goldstein, T., Yorke, J., Fermuller, C. \& Aloimonos, Y. Network Deconvolution. In: {\em International Conference on Learning Representations. 2020.}
\bibitem{saxe2014exact}
Saxe, A., , J. \& Ganguli, S. Exact solutions to the nonlinear dynamics of learning in deep linear neural networks.  In: {\em Proceedings of the International Conference on Learning Represenatations 2014.}
\end{thebibliography}
\end{document}